\documentclass{article}
\usepackage{amssymb}
\usepackage{ijcai25}

\usepackage{times}
\usepackage{soul}
\usepackage{url}
\usepackage[hidelinks]{hyperref}
\usepackage[utf8]{inputenc}
\usepackage[small]{caption}
\usepackage{graphicx}
\usepackage{amsmath}
\usepackage{amsthm}
\usepackage{booktabs}
\usepackage[switch]{lineno}
\usepackage{multirow}
\usepackage{algpseudocode}
\usepackage[ruled,vlined]{algorithm2e}
\usepackage{xcolor}
\newtheorem{theorem}{Theorem}

\title{EyeSeg: An Uncertainty-Aware Eye Segmentation Framework for AR/VR}

\author{
Zhengyuan Peng$^{1^*}$\and
Jianqing Xu$^{2*\P}$\and
Shen Li$^{3^\P}$\and
Jiazhen Ji$^2$\and
Yuge Huang$^2$\\
Jingyun Zhang$^2$\and
Jinmin Li$^4$\and
Shouhong Ding$^2$\and
Rizen Guo$^2$\and
Xin Tan$^{5}$\And
Lizhuang Ma$^{1,5}$
\affiliations
$^1$Shanghai Jiao Tong University~\\
$^2$Tencent~\\
$^3$National University of Singapore\\
$^4$Tsinghua University~\\
$^5$East China Normal University\\
}

\begin{document}

\maketitle

\let\thefootnote\relax
\footnotetext{$^*$ Equal Contribution}
\footnotetext{$^\P$ Project Leads} 

\begin{abstract}
    Human-machine interaction through augmented reality (AR) and virtual reality (VR) is increasingly prevalent, requiring accurate and efficient gaze estimation which hinges on the accuracy of eye segmentation to enable smooth user experiences. We introduce EyeSeg, a novel eye segmentation framework designed to overcome key challenges that existing approaches struggle with: motion blur, eyelid occlusion, and train-test domain gaps. In these situations, existing models struggle to extract robust features, leading to suboptimal performance. Noting that these challenges can be generally quantified by uncertainty, we design EyeSeg as an uncertainty-aware eye segmentation framework for AR/VR wherein we explicitly model the uncertainties by performing Bayesian uncertainty learning of a posterior under the closed set prior. Theoretically, we prove that a statistic of the learned posterior indicates segmentation uncertainty levels and empirically outperforms existing methods in downstream tasks, such as gaze estimation. EyeSeg outputs an uncertainty score and the segmentation result, weighting and fusing multiple gaze estimates for robustness, which proves to be effective especially under motion blur, eyelid occlusion and cross-domain challenges. Moreover, empirical results suggest that EyeSeg achieves segmentation improvements of MIoU, E1, F1, and ACC surpassing previous approaches. The code is publicly available at \href{https://github.com/JethroPeng/EyeSeg}{https://github.com/JethroPeng/EyeSeg}.\par
\end{abstract}

\begin{figure*}[t]
    \centering
    \includegraphics[scale=0.6]{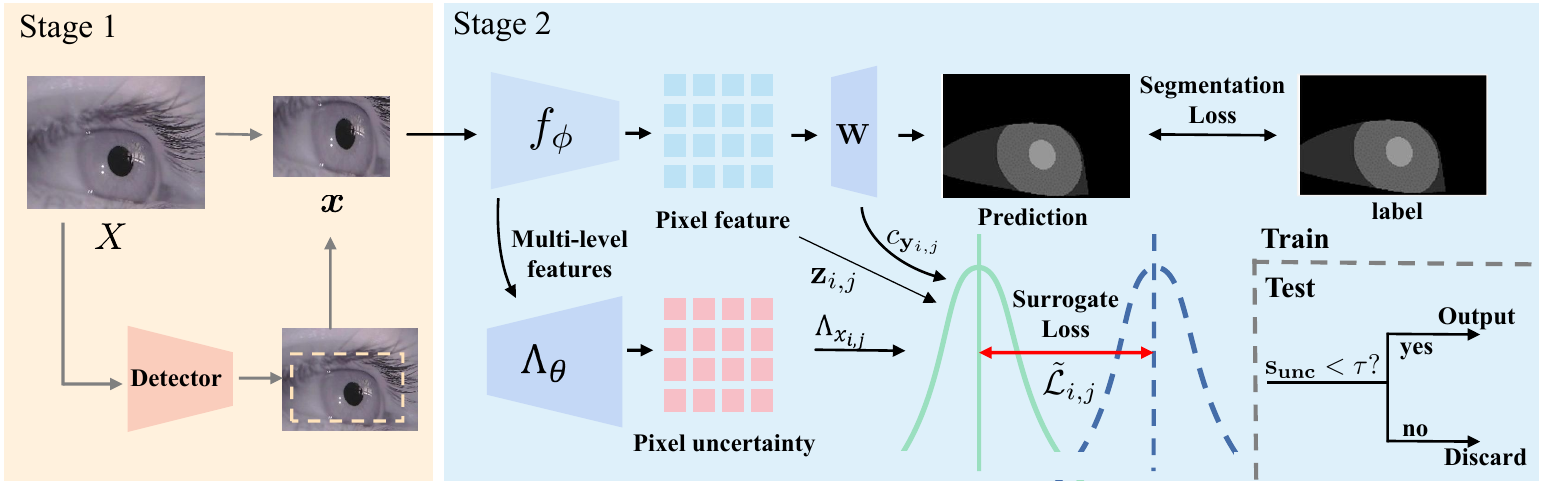}
    \caption{\textbf{The procedure of the proposed framework, EyeSeg.} EyeSeg first detects eye regions, then extracts pixel-wise visual features and uncertainties used for Bayesian learning of a posterior which further yields an uncertainty score $s_{\text{unc}}$ during test. Here, $\tau$ denotes the threshold for decision-making. A detailed description of the process is provided in Algorithm ~\ref{alg:train_alg} and Algorithm ~\ref{alg:test_alg}.}
    \label{fig:framework}
    \vspace{-0.2in}
\end{figure*}  

\section{Introduction}

In recent years, human-machine interaction, especially augmented reality (AR) and virtual reality (VR), is increasingly being popularized by major smart hardware products. For such products to deliver smooth interaction, it is essential to have a gaze estimation algorithm to capture users' iris location and movement accurately and efficiently. Specifically, the algorithm takes as input a real-time image captured by a camera installed in the equipment and returns the resulting segmentation image that specifies the location of the iris in a pixel-wise manner. This segmentation-based product solution usually confronts three challenges: (i) imagery blurriness: due to eye fast motion, images captured usually exhibit blurriness, which poses difficulty to accurate eye segmentation; (ii) eyelid occlusion; (iii) train-test domain gaps: users may use the equipment in various device configurations or varying usage habits, hence inevitably there exists domain gaps between training data and test data. We demonstrate these challenges in Fig.~\ref{fig_show}.

Humans typically express their uncertainty when faced with blurry information or when there are biases in domain knowledge; this allows them to make improvements or reject uncertain conclusions in subsequent analyzes. Inspired by this human trait, we propose a novel uncertainty-aware framework for eye segmentation (EyeSeg) designed to resolve the aforementioned challenges. Unlike previous methods~\cite{wei2021toward,czolbe2021segmentation,sirohi2023uncertainty,fuhl2024pistol} that rely on indirect or proxy measures of uncertainty, our approach explicitly models the uncertainty of the segmentation results. Specifically, we address the variations of input iris images within the eye regions which form a closed set hereinafter. Compared to open-set, the nature of closed-set conveniently induces a desired prior which a posterior approximates to achieve Bayesian uncertainty learning. Leveraging this prior, we utilize a statistic of the learned posterior to estimate posterior probabilities. Theoretically, we prove that the statistic of the learned posterior reflects the level of uncertainty for segmentation results.\par

In our framework, beyond the conventional segmentation pipeline, we extract multi-level features from the segmentation network and predict per-pixel variance, which is then used to compute the overall uncertainty of the image. During the testing phase, if the uncertainty exceeds a predefined threshold, the corresponding prediction is discarded. The pipeline is shown in Figure~\ref{fig:framework}. This uncertainty score can be used as a weight to yield a more reliable gaze estimate, especially in the conditions of eyelid occlusion, imagery blurriness and in-domain/cross-domain settings. Moreover, our algorithm operates directly on the detected eye patch, leading to significant improvements in metrics such as Mean Intersection over Union (MIoU), E1, F1, and ACC, while maintaining computational efficiency with only 1.53G FLOPs.\par

To sum up, our major contributions are listed Ías follows:\par
$\bullet$
\textbf{(Novelty)} We propose a novel uncertainty-aware eye segmentation framework that resolves common challenges in AR/VR.\par
$\bullet$
\textbf{(Interpretability)} We prove that a statistic of the learned posterior provably reflects the level of uncertainty of segmentation results.\par
$\bullet$
\textbf{(Evaluation)} We conduct extensive experiments on multiple real-world datasets and challenge situations, demonstrating significant improvements in metrics such as MIoU, E1, F1, and ACC with 1.53G FLOPs.\par

\section{Related Work}

\noindent\textbf{Eye Segmentation.}  Semantic segmentation is a fundamental task in computer vision with numerous applications, and there have been significant advancements \cite{long2015fully,OlafRonneberger2015UNetCN,badrinarayanan2017segnet,chen2014semantic,chen2017deeplab,chen2017rethinking,chen2018encoder,gong2021boundary,xie2021segformer,yuan2023full,tan2023positive,peng2023generalized,yuan2023full,erisen2024sernet,chen2024beyond,liu2024primitivenet} in recent years. An important application scenario is gaze estimation in AR/VR, where eye segmentation serves as a crucial preprocessing step. Various approaches\cite{hansen2005eye,swirski2012robust,chaudhary2019ritnet,yiu2019deepvog,feng2022real,luo2020shape,wang2020towards,kothari2021ellseg,biswas2023framework,fuhl2024pistol} have been developed, taking into account both the unique characteristics of eye images and the real-time requirements of such applications. Traditional methods\cite{hansen2005eye,swirski2012robust} typically employ morphological operations and geometric features to extract pupil regions from eye images. Deep learning-based methods\cite{chaudhary2019ritnet,yiu2019deepvog,feng2022real,luo2020shape,wang2020towards,kothari2021ellseg,biswas2023framework}, particularly convolutional neural networks, have shown promising results in eye segmentation. These methods leverage the ability of CNNs to learn hierarchical features from large amounts of data, which can be beneficial in handling complex and challenging scenarios. Some CNNs methods\cite{wang2020towards,kothari2021ellseg,biswas2023framework} also utilize ellipse fitting to obtain the shapes of the iris and pupil. Unlike these methods mentioned above, our work develops an uncertainty-aware eye segmentation framework which provides theoretically-grounded uncertainty score besides eye segmentation results, offering an \emph{efficient}, \emph{accurate}, and \emph{robust} solution for eye segmentation. Recently, transformer-based methods \cite{hassan2022sipformer,wei2021toward} have been proposed for eye segmentation. However, due to their high computational complexity, these methods often fail to meet real-time requirements, making them unsuitable for AR/VR applications. \par

\noindent\textbf{Uncertainty Estimation.} 
Existing methods mainly includes Probabilistic methods \cite{kendall2017uncertainties,czolbe2021segmentation}, Evidence Learning\cite{sirohi2023uncertainty}, Deterministic Methods \cite{czolbe2021segmentation,wei2021toward} and Ensemble methods \cite{czolbe2021segmentation,landgraf2024dudes}. In particular, BiTrans~\cite{wei2021toward} leverages uncertainty to refine the eye segmentation, but it is not designed for eye segmentation in VR/AR scenes. Our approach belongs to the category of probabilistic methods and exhibits a high degree of interpretability.    In contrast to other probabilistic methods, our approach directly models uncertainty in a closed set, where uncertainty is computed only at the subregion level. We leverage uncertainty estimation to reject erroneous outputs, instead of using it solely for improving model performance.\par

\section{Methodology}

Our proposed approach, EyeSeg, first detects eye regions, followed by uncertainty-aware eye segmentation. Eye regions provide a closed set, which conveniently induces a desired prior for posterior approximation to achieve uncertainty awareness. The entire approach is illustrated in Fig.~\ref{fig:framework}.

\noindent\textbf{Notation.}
Throughout the paper, we let $(X, Y)$ be a training pair drawn from a data set, where $X$ denote the raw gray-scale training image and $Y$ the corresponding $K$-class pixel-level label (i.e. the segmentation map). Note that $X$ and $Y$ are two-dimensional tensors of which $X_{i,j}$ represents the pixel value and $Y_{i,j}$ denotes the segmentation label at position $(i, j)$. Since $X$'s are captured by a head camera, $X$'s contain inference background besides eyes. Hence, $Y_{i,j}$'s reside in an open set that may contain unknown classes.

\subsection{Eye Detection}
In this step, we aim to exclude interference regions from the image, which typically contains uncontrolled noisy backgrounds. To achieve this, we use object detection techniques to localize eye sockets. By accurately locating the eye socket, we can effectively identify the crucial subregions for eye segmentation. This localization step allows us to narrow down the search space and achieve more efficient and accurate eye segmentation.

Specifically, we first train an eye detection model $d(\cdot)$ that takes $X$ as input and returns a rectangular bounding box $(l, t, h, w)$ that envelops a subregion $\boldsymbol{x}$ which only contains an eye (see the supplement for the specific training procedure). Using the bounding box, the corresponding segmentation label $Y$ is also cropped into a subregion label $\boldsymbol{y}$. All the resulting $\boldsymbol{x}$ and $\boldsymbol{y}$ are resized into $H\times W$. Note that $\boldsymbol{y}_{i,j}$'s now reside in a closed-set $\{0, 1, 2, 3\}$ that corresponds to background, eye, iris, and pupil, respectively. We denote the distribution of the resulting data pair $(\boldsymbol{x}, \boldsymbol{y})$ as $\mathcal{D}$.

\subsection{Deterministic Eye Segmentation}
We then train a deterministic segmentation network to learn a mapping $f_{\phi}: \boldsymbol{x} \in \mathbb{R}^{H\times W} \mapsto \boldsymbol{z} \in \mathbb{R}^{H\times W \times D}$, followed by a linear transform and softmax activation on $\boldsymbol{z}$ at each location $(i,j)$: 
\begin{equation}
    \boldsymbol{p}_{i,j} = \texttt{softmax}(\boldsymbol{W} \boldsymbol{z}_{i,j}),
\label{eq:softmax}
\end{equation}
where $\boldsymbol{W}$ is a 4-by-$D$ matrix ($D$ is the dimensionality of $\boldsymbol{z}$) and $\boldsymbol{p}_{i,j} = [\boldsymbol{p}_{i,j}^{(0)}, \boldsymbol{p}_{i,j}^{(1)}, \boldsymbol{p}_{i,j}^{(2)}, \boldsymbol{p}_{i,j}^{(3)}]^T$ is a probability vector, of which each component, $\boldsymbol{p}_{i,j}^{(c)}$, suggests the probability that the corresponding pixel $(i,j)$ belongs to the class $c$ (for $c = 0, 1, 2, 3$). The deterministic segmentation network is trained using the following cross-entropy loss:
\begin{equation}
\label{eq:seg}
    \min_{\phi, \boldsymbol{W}}\mathcal{L}_{\text{seg}} := \mathbb{E}_{(\boldsymbol{x}, \boldsymbol{y}) \sim \mathcal{D}}\left[-\sum_{i,j} {\boldsymbol{y}_{i,j}\log \boldsymbol{p}_{i,j}}\right]
\end{equation}
During inference, the deterministic segmentation model outputs a segmentation map $\boldsymbol{\hat{y}} \in \mathbb{R}^{H\times W}$, of which the value at each position $(i,j)$ is determined by
\begin{equation}
    \boldsymbol{\hat{y}}_{i,j} = \arg\max_{c} {\boldsymbol{p}_{i,j}^{(c)}}
\end{equation}

\subsection{Projection Head for Uncertainty-Awareness}
\begin{figure}[]
    \centering
    \includegraphics[scale=0.305]{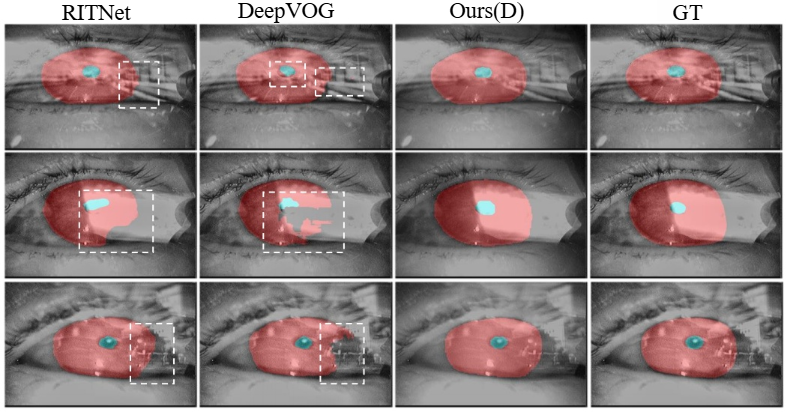}
    \caption{ \textbf{Segmentation Results.} We visually showcase the segmentation results of our approach compared to RITNet and DeepVOG on the Else dataset. The boxes
    represent the errors in segmentation outcomes. It serves as a compelling testament to the outstanding precision and resilience exhibited by our approach}
    \label{fig:show}
\end{figure}

The deterministic segmentation model could not handle imagery blurriness and train-test domain gaps, which we find quite ubiquitous in the context of AR/VR products (see Fig.~\ref{fig:show} for the examples). In light of this, we build an uncertainty-aware projection header on top of the backbone of the pretrained deterministic segmentation model. This header takes the feature maps of each stage of the segmentation network as input and predicts the covariance of each pixel, thus providing an estimate of the uncertainty associated with the model's predictions. The final output is the uncertainty of an entire image calculated from the pixel-level covariance. 

To achieve this, we resort to probabilistic machinery by transforming the deterministic eye segmentation into a probabilistic one. Specifically, we assume that the latent code of a given pixel is no longer a deterministic one but a random variable that follows a Gaussian distribution: $ p_{\theta}(\boldsymbol{z}_{i,j} | \boldsymbol{x}) = \mathcal{N}(\boldsymbol{z}_{i,j}; f_{\phi}(\boldsymbol{x}, i, j), \Lambda_{\theta}({\boldsymbol{x}, i, j}))$, 
where $\Lambda_{\theta}(\cdot)$ is the uncertainty-aware projection head that takes $\boldsymbol{x}$ and an image position $(i,j)$ as input and outputs a diagonal matrix $\Lambda_{\boldsymbol{x}_{i,j}} = \Lambda_{\theta}(\boldsymbol{x}, i, j)$, which captures uncertainty in the latent space. Note that this projection head can be instantiated by any network with learnable parameters $\theta$. 

To learn this posterior distribution $p_{\theta}(\boldsymbol{z}_{i,j} | \boldsymbol{x}, i, j)$, we notice that the matrix $\boldsymbol{W}$ pretrained in Eq.~\eqref{eq:softmax} contains prior knowledge to be utilized: the four rows of $\boldsymbol{W}$ are ideal class template vectors for background, eye, iris, and pupil, respectively. This induces a desired prior $q$ to which $p_{\theta}$ should be regularised. Specifically, the prior $q$ is chosen to be the Dirac delta function, that is, $q(\boldsymbol{z}_{i,j}|\boldsymbol{y}_{i,j}) = \delta(\boldsymbol{z}_{i,j} - \boldsymbol{c}_{\boldsymbol{y}_{i,j}})$, where $\boldsymbol{c}_{\boldsymbol{y}_{i,j}} := \boldsymbol{W}^T\boldsymbol{e}_{\boldsymbol{y}_{i,j}}$. Here the vector $\boldsymbol{e}_{\boldsymbol{y}_{i,j}}$ is a one-hot vector which contains one at the index $\boldsymbol{y}_{i,j}$ and zeros at the rest of indices. Intuitively, the matrix-vector multiplication $\boldsymbol{W}^T\boldsymbol{e}_{\boldsymbol{y}_{i,j}}$ takes the ${\boldsymbol{y}_{i,j}}^{\text{th}}$ row of $\boldsymbol{W}$ and transposes it as a column vector.

To perform regularization, we minimize the cross entropy (CE) between the chosen prior $q$ and the posterior $p_{\theta}$ at all image positions $(i=1,...,H, j=1,..., W)$:

\begin{equation}
\label{orloss}
\small
\begin{split}
  & \min_{\theta}   \mathbb{E}_{(\boldsymbol{x}, \boldsymbol{y}) \sim \mathcal{D}} \left[ \sum_{i,j}\underbrace{\operatorname{CE} ( q(\boldsymbol{z}_{i,j}|\boldsymbol{y}_{i,j})~||~p_{\theta}(\boldsymbol{z}_{i,j} | \boldsymbol{x}_{i, j}))}_{:=\mathcal{L}_{i,j}} \right] \\
\end{split}
\end{equation}
where each summand $\mathcal{L}_{i,j} =$
\begin{equation}
\label{eq:summand}
\small
    {\frac{1}{2}(\boldsymbol{c}_{\boldsymbol{y}_{i,j}}-\boldsymbol{z}_{i,j} )^{T} \Lambda_{\boldsymbol{x}_{i,j}}^{-1} ( \boldsymbol{c}_{\boldsymbol{y}_{i,j}}-\boldsymbol{z}_{i,j} ) + \frac{\ln \operatorname{det}(\Lambda_{\boldsymbol{x}_{i,j}})}{2} + \frac{D}{2} \ln 2 \pi}
\end{equation}

Here, Eq.~\eqref{eq:summand} is obtained by plugging $q(\boldsymbol{z}_{i,j}|\boldsymbol{y}_{i,j}) = \delta(\boldsymbol{z}_{i,j} - \boldsymbol{c}_{\boldsymbol{y}_{i,j}})$ and $ p_{\theta}(\boldsymbol{z}_{i,j} | \boldsymbol{x}) = \mathcal{N}(\boldsymbol{z}_{i,j}; f_{\phi}(\boldsymbol{x}, i, j), \Lambda_{\theta}({\boldsymbol{x}, i, j}))$ into Eq.~\eqref{orloss}. 

After optimization, we obtain the optimal $\theta^*$ and further, given a test $\boldsymbol{\Tilde{x}}$, its optimal $\Lambda_{\boldsymbol{\Tilde{x}_{i,j}}}^*$ that captures the dimension-wise latent uncertainty at the pixel location $(i,j)$ of $\boldsymbol{\Tilde{x}}$. However, we may expect one summary scalar that quantifies the uncertainty at $(i,j)$. Next, we theoretically show that the desired scalar happens to be the trace of the optimal $\Lambda_{\boldsymbol{x}_{i,j}}^*$.

\begin{theorem}
\label{theorem1}
Given an image $\boldsymbol{x}$ and a position $(i,j)$ of interest, the trace of the optimal $\Lambda_{\boldsymbol{x}_{i,j}}^*$ that minimizes each summand of the proposed loss function Eq.~\eqref{orloss} is equal to the squared Euclidean distance between the latent code $\boldsymbol{z}_{i,j}$ and the corresponding class center $\boldsymbol{c}_{\boldsymbol{y}_{i,j}}$:
\begin{equation}\label{tr}
    \operatorname{tr}(\Lambda_{\boldsymbol{x}_{i,j}}^*) = ||\boldsymbol{z}_{i,j} - \boldsymbol{c}_{\boldsymbol{y}_{i,j}}||_2^2
\end{equation}
where 
\begin{equation}
\begin{split}
\Lambda_{\boldsymbol{x}_{i,j}}^* = \operatorname{diag}(\sigma_1^2, ..., \sigma_D^2)
&= \arg\min_{\Lambda_{\boldsymbol{x}_{i,j}}}{\mathcal{L}_{i,j}}
\end{split}
\end{equation}
\end{theorem}

\begin{algorithm}[]
    \caption{Training Algorithm}
    \label{alg:train_alg}
    \KwIn{The training pairs $(X, Y)\sim \mathcal{D}$; The pretrained detection model $d(\cdot)$.}
    \KwOut{The deterministic segmentation network $f_{\phi}$, the matrix $\boldsymbol{W}$, the projection head $\Lambda_\theta$} 
    
    \For{$(X,Y)\sim \mathcal{D}$}{
        $(l,t,h,w) \gets d(X)$;\\
        $\boldsymbol{x} \gets$ crop $X$ using $(l,t,h,w)$;\\
        $\boldsymbol{y} \gets$ crop $Y$ using $(l,t,h,w)$;
    }
    
    $\{\phi$, $\boldsymbol{W}\}$ $\gets$ Train the eye segmentation network by minimizing $\mathcal{L}_{\text{seg}}$ in Eq.~\eqref{eq:seg};

    $\{\theta\}$ $\gets$ Train the projection head by minimizing the surrogate loss in Eq.~\eqref{uploss};
    
    \Return $f_{\phi}$, $\boldsymbol{W}$, $\Lambda_\theta$
    
\end{algorithm}

Theorem~\ref{theorem1} suggests that the closer the latent code $\boldsymbol{z}_{i,j}$ gets to $\boldsymbol{c}_{\boldsymbol{y}_{i,j}}$, the lower the value of $\operatorname{tr}(\Lambda_{\boldsymbol{x}_{i,j}}^*)$ becomes. The Euclidean distance can be interpreted as the distance of the prediction at position $(i,j)$ towards its corresponding ideal position in the latent space, and therefore the quantity $\operatorname{tr}(\Lambda_{\boldsymbol{x}_{i,j}}^*)$ can be seen as the uncertainty for the prediction. In the inference stage, the category of a pixel cannot be obtained in advance, so predicting the category center to which the pixel at $(i,j)$ belongs is an ill-conditioned problem. Our method circumvents this difficulty by estimating an uncertainty $\operatorname{tr}(\Lambda_{\boldsymbol{x}_{i,j}}^*)$ that mathematically measures how close a pixel under test is to its unknown class center.

\subsection{Optimization Analysis}
Solving the original loss function Eq.~\eqref{orloss} requires minimizing each summand of Eq.~\eqref{orloss} simultaneously. In practice, we find it often converges to suboptimal minima. In this section, we provide a detailed analysis and propose a surrogate loss that can circumvent this optimization difficulty by virtue of Theorem~\ref{theorem1}.

\begin{figure}[t]
    \centering
    \includegraphics[scale=0.251]{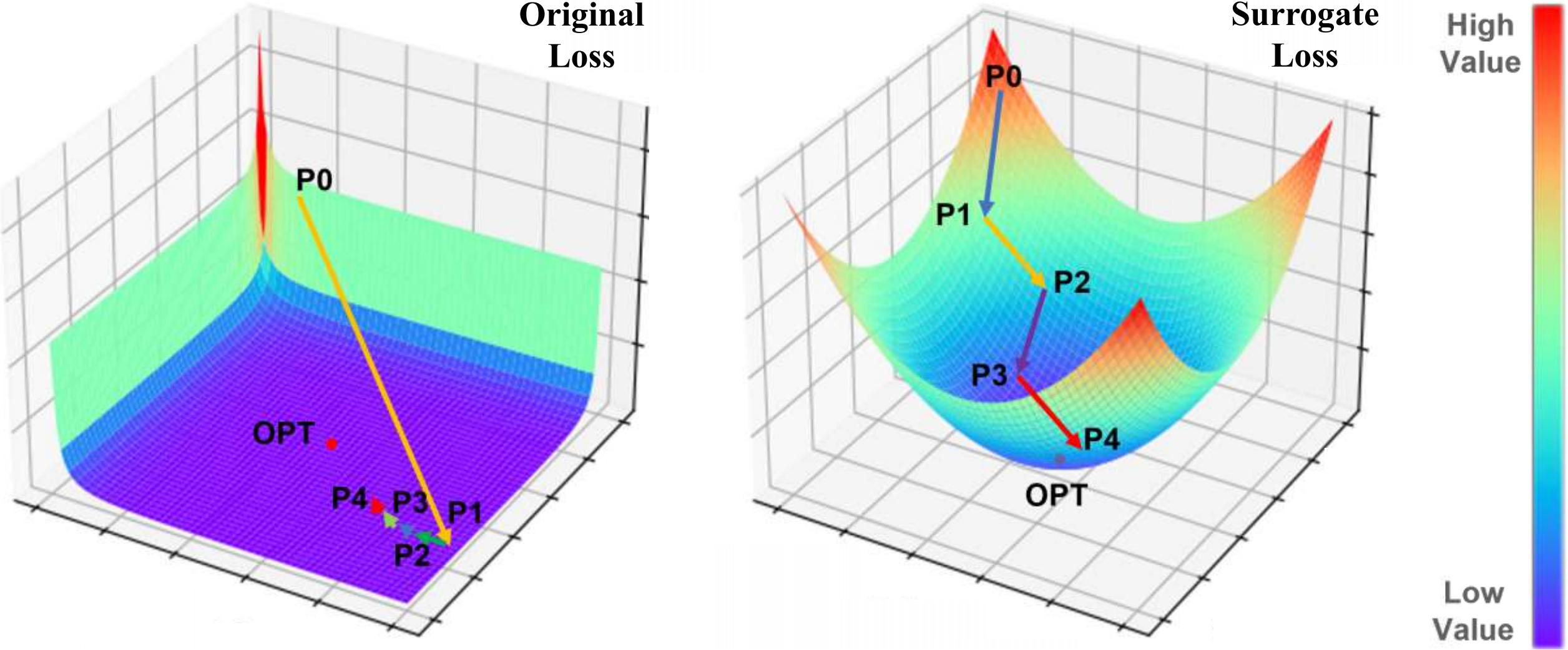}
    \caption{\textbf{The function landscapes of the original loss Eq.~\eqref{orloss} (\textbf{left}) and the surrogate loss Eq.~\eqref{uploss} (\textbf{right})}. For visualization, these loss functions are illustrated as functions of two free variables $\text{w}1, \text{w}2$ of the covariance $\Lambda_{\boldsymbol{x}_{i,j}}$. We mark an optimization path (\textbf{P0} $\rightarrow $ \textbf{P4}) and the optimal point (\textbf{OPT}) for each. It can be observed that the original loss is prone to optimization issues such as gradient vanishing, whereas the surrogate loss is easy to optimize.}
    \label{fig:landscape}
    \vspace{-0.2in}
\end{figure}

From Eq.~\eqref{orloss}, we can find the first derivative of the loss function $\mathcal{L}_{i,j}$: 

\begin{equation}
      \frac {\partial \mathcal{L}_{i,j} }{\partial \Lambda_{\boldsymbol{x}_{i,j}}} 
      = \frac{1}{ 2} \Lambda_{\boldsymbol{x}_{i,j}}^{-T} \left( I - (\boldsymbol{c}_{\boldsymbol{y}_{i,j}}-\boldsymbol{z}_{i,j}) (\boldsymbol{c}_{\boldsymbol{y}_{i,j}}-\boldsymbol{z}_{i,j})^T \Lambda_{\boldsymbol{x}_{i,j}}^{-T}\right) 
\end{equation}
There are two multiplicative factors that contribute to decreasing the loss derivative. The first factor leads to a trivial solution where the entries of $\Lambda_{\boldsymbol{x}_{i,j}}$ are extremely large. In practice, we observe that the first factor dominates in finding the minimum of Eq.~\eqref{orloss} due to random initialization of network parameters in the optimization process. This means that the entire optimization process will probably face optimization issues such as gradient vanishing. In Fig.~\ref{fig:landscape} (left) we showcase the function landscape of the loss Eq.~\eqref{orloss} as a function of $\Sigma_{\boldsymbol{x}_{i,j}}$ in the two-dimensional case.

To circumvent the optimization difficulty of loss Eq.~\eqref{orloss}, we derive a surrogate loss based on Theorem~\ref{theorem1}. According to Theorem~\ref{theorem1}, the diagonals are all that we need for uncertainty estimation. Therefore, we only have to have the uncertainty module $\Lambda_{\theta}(\cdot)$ to output the diagonals and make them approximate the desired quantity that is only available during training, $(\boldsymbol{c}_{\boldsymbol{y}_{i,j}}-\boldsymbol{z}_{i,j}) \odot (\boldsymbol{c}_{\boldsymbol{y}_{i,j}}-\boldsymbol{z}_{i,j})$:

\begin{equation}
    \min_{\theta} \mathbb{E}_{(\boldsymbol{x}, \boldsymbol{y}) \sim 
      \mathcal{D}} \left[ \sum_{i,j} \Tilde{\mathcal{L}}_{i,j}\right]
      \label{uploss}
\end{equation}

where each summand $\Tilde{\mathcal{L}}_{i,j}$ is
\begin{equation}
    \Tilde{\mathcal{L}}_{i,j} = \left\lVert \operatorname{diag}(\Lambda_{\boldsymbol{x}_{i,j}}) - (\boldsymbol{c}_{\boldsymbol{y}_{i,j}}-\boldsymbol{z}_{i,j}) \odot (\boldsymbol{c}_{\boldsymbol{y}_{i,j}}-\boldsymbol{z}_{i,j}) \right\rVert_2^2
\end{equation}

Note that the function landscape of the surrogate loss Eq.~\eqref{uploss} is flat and smooth as compared to the original loss Eq.~\eqref{orloss}, as shown in Fig.~\ref{fig:landscape}. Therefore, the optimization objective is more amenable, as will be shown in our experiments. See Algorithm~\ref{alg:train_alg} for the detailed training procedure.

\subsection{Inference}
After training, we obtain the resulting eye segmentation model $f_{\phi}$, the pretrained matrix $\boldsymbol{W}$ and the projection head $\Lambda_\theta$. Then, given a raw testing image $X$, we first use the pretrained eye detection model $d(\cdot)$ to obtain the patch $\boldsymbol{x}$, then use $f_{\phi}$ and $\boldsymbol{W}$ to obtain the segmentation result $\hat{\boldsymbol{y}}$. Further, besides the segmentation result, we can obtain the predicted pixel-wise covariance $\Lambda_{\boldsymbol{x}_{i,j}} = \Lambda_\theta(\boldsymbol{x}, i, j)$ at every position $(i,j)$. However, we expect an associated uncertainty score $s_{\text{unc}}$ that can summarize the uncertainty of the whole predicted result $\hat{\boldsymbol{y}}$. In this section, we show how to derive this quantity from the predicted pixel-wise covariance.

\begin{algorithm}[]
    \caption{Test Algorithm}
    \label{alg:test_alg}
    \KwIn{The testing image $X$; The pretrained eye detection model $d(\cdot)$; The eye segmentation model $f_{\phi}$, the pretrained matrix $\boldsymbol{W}$, the projection head $\Sigma_\theta$.}
    \KwOut{The predicted segmentation map $\boldsymbol{\hat{y}}$ and its associated uncertainty score $s_{\text{unc}}$} 
    
    $(l,t,h,w) \gets d(X)$;\\
    $\boldsymbol{x} \gets$ crop $X$ using $(l,t,h,w)$;\\
    
    $\boldsymbol{z} \gets f_{\phi}(\boldsymbol{x})$;\\

     \For{$(i,j)$}{
        $\boldsymbol{p}_{i,j} \gets \texttt{softmax}(\boldsymbol{W}\boldsymbol{z}_{i,j})$;\\
        $\boldsymbol{\hat{y}}_{i,j} \gets \arg\max_{c} {\boldsymbol{p}_{i,j}^{(c)}}$; \\
        $\Lambda_{\boldsymbol{x}_{i,j}} \gets \Lambda_{\theta}(\boldsymbol{x}, i, j)$ \\
    }
    
    $s_{\text{unc}} \gets \sum_{i,j} \ln {\operatorname{det}(\Lambda_{\boldsymbol{x}_{i,j}})}$
    
    \Return $\boldsymbol{\hat{y}}$, $s_{\text{unc}}$
\end{algorithm}

We propose to use the negative logarithm of the posterior density $p(\boldsymbol{z} | \boldsymbol{x})$ as the uncertainty score $s_{\text{unc}}$. The intuition is that the higher the density $p(\boldsymbol{z} | \boldsymbol{x})$ is, the lower the uncertainty score. Specifically, the posterior density can be factorized into:\par
\begin{equation}
\small
    p(\boldsymbol{z}|\boldsymbol{x}) 
    = \prod_{i,j} p(\boldsymbol{z}_{i,j}|\boldsymbol{x})
    = \prod_{i,j} \frac{e^{-\frac{1}{2} ( \boldsymbol{c}_{\boldsymbol{y}_{i,j}}-\boldsymbol{z}_{i,j} )^{T} \Lambda_{\boldsymbol{x}_{i,j}}^{*-1} ( \boldsymbol{c}_{\boldsymbol{y}_{i,j}}-\boldsymbol{z}_{i,j} )}}{\sqrt{2\pi \operatorname{det}(\Lambda_{\boldsymbol{x}_{i,j}}^*) }} 
\label{posterior}
\end{equation}

Let $\boldsymbol{v} = (\boldsymbol{c}_{\boldsymbol{y}_{i,j}} - \boldsymbol{z}_{i,j})$. Then, for the quadratic form
\begin{align}
\gamma_{i,j} = (\boldsymbol{c}_{\boldsymbol{y}_{i,j}} - \boldsymbol{z}_{i,j})^{T} \Lambda_{\boldsymbol{x}_{i,j}}^{*-1} (\boldsymbol{c}_{\boldsymbol{y}_{i,j}} - \boldsymbol{z}_{i,j}),
\end{align}

\begin{align}
    \operatorname{tr}(\gamma_{i,j}) &= \operatorname{tr}\left(\boldsymbol{v}^{T} 
        \begin{pmatrix}
            \sigma_1^{-2} &        &        \\
                            & \ddots &        \\
                            &        & \sigma_D^{-2}
        \end{pmatrix}
        \boldsymbol{v}\right) \\[1.5ex]
    &= \operatorname{tr}\left(\boldsymbol{v}^{T} 
        \left(
            \begin{smallmatrix}
                \frac{1}{\boldsymbol{v}_1^2} &        &        \\
                                 & \ddots &        \\
                                 &        & \frac{1}{\boldsymbol{v}_D^2}
            \end{smallmatrix}
        \right)
        \boldsymbol{v}\right) \\[1.5ex]
    &= \operatorname{tr}\left(\boldsymbol{v}\boldsymbol{v}^{T}
            \left(
                \begin{smallmatrix}
                    \frac{1}{\boldsymbol{v}_1^2} &        &        \\
                                     & \ddots &        \\
                                     &        & \frac{1}{\boldsymbol{v}_D^2}
                \end{smallmatrix}
            \right)
            \right) \\[1.5ex]
    &= \operatorname{tr}\left( 
            \left(
                \begin{smallmatrix}
                    \boldsymbol{v}_1^2 &    \boldsymbol{v}_1\boldsymbol{v}_2    &        \\
                    \boldsymbol{v}_2\boldsymbol{v}_1     & \ddots &        \\
                         &        & \boldsymbol{v}_D^2
                \end{smallmatrix}
            \right)
            \left(
                \begin{smallmatrix}
                    \frac{1}{\boldsymbol{v}_1^2} &        &        \\
                                     & \ddots &        \\
                                     &        & \frac{1}{\boldsymbol{v}_D^2}
                \end{smallmatrix}
            \right)
            \right) \\[1.5ex]
   &= \operatorname{tr}\left(  \left(\begin{array}{cccc}
        1 & * & \cdots & * \\
        * & 1 & \cdots & * \\
        \vdots & \vdots & \ddots & \vdots \\
        * & * & \cdots & 1
        \end{array} \right) \right) = D
\label{cz}
\end{align}

Substituting Eq.~\eqref{cz} into Eq.~\eqref{posterior}. we can derive the uncertainty estimate for the whole result from pixel-level estimate:
\begin{equation}
    p(\boldsymbol{z}|\boldsymbol{x}) \propto \prod_{i,j} \frac{1}{\sqrt{\operatorname{det}(\Lambda_{\boldsymbol{x}_{i,j}}^*) }}
\end{equation}
Therefore, our proposed uncertainty score $s_\text{unc}$ for the whole segmented result is given by
\begin{equation}
    - \ln p(\boldsymbol{z}|\boldsymbol{x}) \propto ~ \sum_{i,j} \ln \operatorname{det}( \Lambda_{\boldsymbol{x}_{i,j}}^*) = s_{\text{unc}}
\end{equation}

The uncertainty score $s_\text{unc}$ quantifies the predictive confidence of the model in its segmentation output by mapping the posterior probability to the real number domain
$\mathbb{R}$. See Algorithm~\ref{alg:test_alg} for the procedure of test algorithm.

\begin{table*}[t]
\centering
\small
\begin{tabular}{l | c | c | c | c | c | c | c | c| c |c}
\toprule[1.5pt]
\hline
\multirow{2}{*}{Methods} & \multicolumn{2}{c|}{Else} & \multicolumn{2}{c|}{Dikablis} & \multicolumn{2}{c|}{LPW} & \multicolumn{2}{c|}{OpenEDS} &  \multirow{2}{*}{Average} &  \multirow{2}{*}{Model FLOPs} \\
\cline{2-9}
& Iris & Pupil & Iris & Pupil & Iris & Pupil & Iris & Pupil & &\\
\hline
DeepVOG~\cite{yiu2019deepvog} & 93.0 & 87.0 & 91.1 & 94.2 & 79.7 & 83.9 & N/A & 89.1&N/A & 3.5G\\ 
RITnets~\cite{chaudhary2019ritnet} & 93.7 & 88.7 & 91.6 &  94.7 & 86.7 & 86.3 & 91.4 & 95.0 &91.0 & 16.57G\\ 
Pylids~\cite{biswas2023framework} &88.2&82.3&89.8&87.2&85.9&89.0&85.7&88.4&87.1&21.15G\\
Pisto*~\cite{fuhl2024pistol} &89.8&83.3& N/A & N/A & N/A& N/A&88.6&87.0&N/A &N/A\\
DeepLabv3+~\cite{chen2018encoder} & 93.7 &84.4&92.2&94.0&85.7&93.8&98.2&96.7& 92.3&39.41G\\ 
Ours(D) & 95.1 & 87.6 & 91.7 & 94.2& 90.1 & 93.5 & 98.3 & 97.3 & 93.5 & \textbf{125.85M}\\ 
Ours(E) & \textbf{96.0} & \textbf{90.4} & \textbf{93.2} & \textbf{95.1} & \textbf{91.9} & \textbf{95.2} & \textbf{98.6} & \textbf{97.8} & \textbf{94.8} & 1.53G\\
\bottomrule[1.5pt]
\hline
\end{tabular}
\caption{Segmentation Results (MIoU) on Various Challenging Benchmarks. Best results are marked in bold. Methods marked with * use the application from the papers due to conditions, without task-specific retraining.}
\label{tab_combined}
\begin{center}
\end{center}
\end{table*}

\section{Experiments}
\subsection{Datasets}

There are several publicly available eye segmentation datasets \cite{fuhl2015excuse,fuhl2016else,garbin2020dataset,fuhl2021teyed,tonsen2016labelled,fuhl2016pupilnet,fuhl2017pupilnet} that have been established to aid research and algorithm development. We conduct extensive experiments to evaluate our proposed EyeSeg on multiple widely used datasets, including OpenEDS~\cite{openeds}, LPW~\cite{LPW}, Dikablis~\cite{fuhl2022teyed}. These datasets provide a diverse range of real-world scenarios and variations in imaging conditions. Furthermore, we also curate a meticulously annotated dataset, Else~\cite{fuhl2016else} to assess the performance of competing methods under more challenging conditions. The annotated labels will be open-sourced. All datasets include four classes: background, eye, iris, and pupil.\par

\subsection{Implementation Details}

To increase the diversity of training data, we use conventional data augmentation techniques such as random rotation, translation, scaling, and horizontal flipping. All experimental settings are kept the same for fair comparison. We apply gamma correction to the image during the preprocessing stage.\par

In the preprocessing step, to localize and identify the regions of eyes in images, we employ an object detection of which its network architecture is similar to YOLOv3~\cite{yolov3}. Specifically, we employ a pruned, scaled-down version to reduce computational complexity. The network architecture is simplified into a shallower model, with the total computation reduced to 43.13M FLOPs. After this step, all regions of eyes are transformed into $96\times96$ images to serve as input to the segmentation model.\par

We utilize the lightweight segmentation network architectures DeepVOG~\cite{yiu2019deepvog} and DenseElNet~\cite{kothari2021ellseg} as the backbones for segmentation, which we refer to as Ours(D) and Ours(E), respectively. The input size of the network is standardized to $96\times96$, and the output category of network segmentation is set to a closed set including $4$ categories (Pupil, Iris, Eye and Background). Due to eye patch extraction, our framework has a much smaller computational overhead than that of using full-images as input under the same segmentation network architecture.

Our uncertainty-aware projection header is implemented using a bottleneck architecture. Specifically, it consists of one downsampling layer and two upsampling layers with skip connections that link corresponding outputs at each layer. By incorporating these skip connections, the projection header can leverage the rich feature representations extracted by the backbone network. The downsampling layer reduces the spatial dimensions of the input, capturing high-level contextual information. The subsequent upsampling layers recover the spatial resolution, enabling fine-grained uncertainty estimation at each pixel. This architecture design ensures that our uncertainty-aware projection header can effectively capture both global and local information from the backbone network, which contributes to the robustness and reliability of uncertainty estimation.\par

\subsection{Baseline Methods}
In experiments, we consider two experimental settings: segmentation and uncertainty estimation. In segmentation, our proposed model is compared with existing top-performing and lightweight  eye segmentation methods for AR/VR. We consider ad-hoc eye segmentation methods, including DeepVOG~\cite{yiu2019deepvog}, RITNet~\cite{chaudhary2019ritnet},
Pisto~\cite{fuhl2024pistol} and Pylids~\cite{biswas2023framework}, with settings consistent with the original training settings in their respective papers. 
Pylids~\cite{biswas2023framework} is a model for keypoint detection and ellipse fitting. Pisto~\cite{fuhl2024pistol} executes the executable program provided in the paper. For fair comparison, only results from test datasets not used in training are evaluated. In addition, we also consider a generic segmentation method DeepLabv3+~\cite{chen2018encoder} using MobileNet backbone. Notably, it has significantly higher computational complexity than the above methods.\par

In uncertainty estimation, baseline methods include Ensemble~\cite{czolbe2021segmentation}, EvPSNet~\cite{sirohi2023uncertainty}, Dudes~\cite{landgraf2024dudes} and BiTrans~\cite{wei2021toward}. To ensure a fair comparison, we use DeepVOG as the backbone for all the methods.\par

\subsection{Results}
In this section, we demonstrate EyeSeg's performance in terms of two experimental settings: segmentation and uncertainty estimation. 
Performance is measured using metrics including MIoU (Mean Intersection over Union), E1, F1 and ACC. See complete results in Appendix.

\textbf{Segmentation.}
Tab.~\ref{tab_combined} shows the segmentation results across four datasets. On the OpenEDS dataset, our approach outperforms the state-of-the-art. This improvement is attributed to our method's ability to exclude irrelevant regions, leading to a more efficient and accurate model. These findings suggest the advantages of EyeSeg in eye semantic segmentation.

\begin{table*}[]
\small
\begin{center}
\begin{tabular}{c | c |c| c| c| c| c| c| c| c| c}
\toprule[1.5pt]
\hline
Settings & \multicolumn{5}{c|}{In-domain} & \multicolumn{5}{c}{Cross-domain}\\
\hline
Thresholds& $1\%$ & $2\%$ & $3\%$ & $4\%$ & $5\%$ & $1\%$ & $2\%$ & $3\%$ & $4\%$ & $5\%$\\
\hline
Ensemble~\cite{czolbe2021segmentation} & 87.11 & 87.14 & 87.21 & 87.26 & 87.33 & 71.49 & 71.62 & 71.73 & 71.84 & 71.93\\
BiTrans~\cite{wei2021toward}  & 86.75 & 86.80 & 86.90 & 86.97 & 87.10 & 75.87 & 75.99 & 76.10 & 76.18 & 76.27\\
EvPSNet~\cite{sirohi2023uncertainty} & 87.28 & 87.28 & 87.45 & 87.48 & 87.48 & 76.61 & 76.73 & 76.79 & 76.86 & 76.90\\
Dudes~\cite{landgraf2024dudes} & 87.65&87.79&87.89&88.01&88.09&74.65&74.74&74.88&74.98&75.06\\
Ours(D)  & \textbf{89.32} & \textbf{89.39} & \textbf{89.46} & \textbf{89.68} & \textbf{89.72} & \textbf{86.56} & \textbf{86.76} & \textbf{86.89} & \textbf{87.02} & \textbf{87.15}\\

\hline

Settings & \multicolumn{5}{c|}{Occlusion} & \multicolumn{5}{c}{Blur}\\
\hline
Thresholds& $1\%$ & $2\%$ & $3\%$ & $4\%$ & $5\%$ & $1\%$ & $2\%$ & $3\%$ & $4\%$ & $5\%$\\
\hline
Ensemble~\cite{czolbe2021segmentation} & 86.50& 86.70& 86.76& 86.93 & 86.93 & 84.53& 84.93 & 84.94& 85.04& 85.06\\
BiTrans~\cite{wei2021toward}   & 86.16& 86.29 & 86.38& 85.48& 86.67& 83.64&83.66 & 83.83& 83.95& 84.20\\
EvPSNet~\cite{sirohi2023uncertainty}  & 86.72 & 86.85& 86.91& 86.91 & 86.89 & 84.13& 84.21& 84.24& 84.25 & 84.20\\
Dudes~\cite{landgraf2024dudes} & 87.21&87.35&87.43&87.56&87.68&84.96&85.10&85.10&85.20&85.21\\
Ours(D)   & \textbf{89.23} & \textbf{89.35} & \textbf{89.53}& \textbf{89.58}& \textbf{89.60} & \textbf{88.04}& \textbf{88.09} & \textbf{88.19}& \textbf{88.29}& \textbf{88.36}\\

\bottomrule[1.5pt]
\hline
\end{tabular}
\caption{Segmentation results (MIoU) after removing images with high uncertainty scores ($s_{\text{unc}}$). Best results are marked in bold. All models are trained and evaluated on the Else dataset, except in the Cross-domain setting, where they are trained on Else and evaluated on OpenEDS.}
\label{tab6}
\end{center}
\end{table*}

\begin{table*}[htp]
\small
\begin{center}
\begin{tabular}{c | c c c c c | c | c c c c c}
\toprule[1.5pt]
\hline
\multirow{2}{*}{Ablation I} & \multicolumn{5}{c|}{Thresholds} & \multirow{2}{*}{Ablation II} & \multicolumn{5}{c}{Thresholds}\\
\cline{2-6} \cline{8-12}
& $1\%$ & $2\%$ & $3\%$ & $4\%$ & $5\%$ && $1\%$ & $2\%$ & $3\%$ & $4\%$ & $5\%$\\
\hline
 Original loss & 86.4 & 86.5 & 86.6 & 86.7 & 86.7 &Entire image & 75.5 & 75.6 & 75.8 & 75.9 & 76.0\\
Surrogate loss & \textbf{86.6} & \textbf{86.8} & \textbf{86.9} & \textbf{87.0} & \textbf{87.2} &Ours(D) &\textbf{86.6} & \textbf{86.8} & \textbf{86.9} & \textbf{87.0} & \textbf{87.2}\\
\bottomrule[1.5pt]
\hline
\end{tabular}
\caption{Ablation study I and II. All models are trained on the Else dataset and evaluated on the OpenEDS dataset.}
\label{tab7}
\end{center}
\end{table*}

\begin{figure}[!t]
    \centering
    \includegraphics[scale=0.46 ]{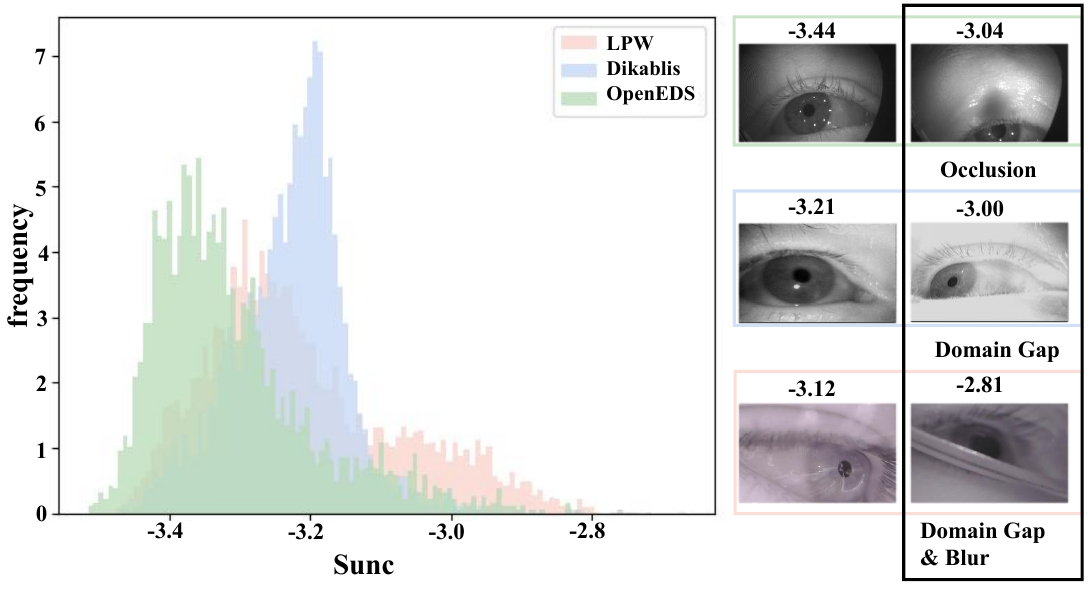}
    \caption{\textbf{Uncertainty distribution of different cross-domain datasets.} 
    The model is trained on the Else dataset and tested on others datasets. Different domains have different distribution modes. Our uncertainty score correlates with eye semantics completeness. Compared to normal images, uncertainty is higher in cases of imagery blurriness, eyelid occlusion, and train-test domain gaps.
    }
    \label{fig_show}
\end{figure}

Additionally, we also evaluate the performance of our method on Else, Dikablis and LPW. It demonstrates that our proposed framework achieves better or promising performance using only a segmentation network with a small amount of computation. In Fig.~\ref{fig_show}, we present a comparison of the segmentation results from our approach against RITNet and DeepVOG on the Else dataset. As shown in the figure, our method demonstrates higher accuracy and robustness across various challenging scenarios. Besides MIoU, we observe that EyeSeg also exhibits superior performance on E1, F1 and ACC in Appendix.

\textbf{Uncertainty Estimation.}
We conduct experiments to evaluate the performance of our uncertainty-aware approach on hard samples in four settings: in-domain, cross-domain, Occlusion and Blur. In the in-domain settings, we evaluate the effects of factors such as image blurriness and noise on model performance. In the cross-domain settings, we evaluate the effects of domain gaps on model performance. In the occlusion setting, we simulate real-world scenarios where parts of the eye are partially covered and assess how well the model can still segment the eye accurately. For the blur setting, we introduce varying levels of motion blur to the test samples to gauge the model’s robustness in maintaining segmentation quality under degraded image conditions.\par 

We collect all images from the dataset and rank them according to uncertainty scores generated by different models. With images of large uncertainty scores filtered out, we expect the rest of images to attain good segmentation results. To ensure fairness, all methods are evaluated based on MIoU. As demonstrated in Tab.~~\ref{tab6}, the images selected by our method show better results than those by other uncertainty-aware methods. This indicates that our proposed uncertainty-aware projection header is more effective for eye segmentation.

In Fig.~\ref{fig_show}, the uncertainty scores obtained by our method align with human cognition: 1) Our proposed uncertainty scores across cross-domain datasets positively correlate with their similarity to the training dataset Else; 2) Within each dataset, the uncertainty scores generally correlate positively with eye semantics completeness.\par

\subsection{Ablation Study}

\noindent\textbf{Study I.} We conduct experimental comparisons of the two optimization objectives proposed Eq.~\eqref{uploss} in this paper. From the results presented in the Tab.~\ref{tab7}, it can be observed that both the optimization objectives yield improvements. The results obtained from each stage of Tab.~\ref{tab7} demonstrate that the surrogate loss function provides a more accurate estimation of the uncertainty associated with the segmentation results.

\noindent\textbf{Study II.} We conduct a comprehensive comparison between our method's uncertainty estimation and segmentation performance on both the entire image and sub-regions. As shown in Tab.~\ref{tab7}, our method consistently outperforms the other regardless of the proportion of data removed according to the uncertainty scores estimated by our framework.

\section{Conclusion}

We have proposed an uncertainty-aware eye segmentation framework, EyeSeg, designed to address major challenge in AR/VR: motion blur, eyelid occlusion and train-test domain gaps. We first extract eye patches and then perform uncertainty estimation and segmentation on them. The extracted eye patches satisfy the closed-set condition, reducing the model’s difficulty in processing redundant background information and enhancing its robustness for cross-domain data segmentation. The patch extraction allows us to perform an uncertainty estimation Bayesian approach under which a posterior approximates a designed prior induced by the closed set. In addition, we introduce a novel theoretical-grounding approach for uncertainty estimation in this task. Our model surpasses the state-of-the-art methods.

\section*{Acknowledgments}
This work is supported by the National Natural Science Foundation of China (62302167, U23A20343, 62472282, 72192821), Young Elite Scientists Sponsorship Program by CAST (YESS20240780),Shanghai Sailing Program (23YF1410500), Chenguang Program of Shanghai Education Development Foundation and Shanghai Municipal Education Commission (23CGA34).

\bibliographystyle{named}
\bibliography{ijcai25}

\end{document}